\documentclass[10pt,twocolumn,letterpaper]{article}
\usepackage{multirow}
\usepackage{verbatim}
\usepackage{siunitx}
\usepackage[pagenumbers]{iccv}

\definecolor{iccvblue}{rgb}{0.21,0.49,0.74}
\usepackage[pagebackref,breaklinks,colorlinks,allcolors=iccvblue]{hyperref}

\def\paperID{3502} 
\def\confName{ICCV}
\def\confYear{2025}

\title{S$^3$CE-Net: Spike-guided Spatiotemporal Semantic Coupling and Expansion Network for Long Sequence Event Re-Identification}

\author{ Xianheng Ma \and
Hongchen Tan \and
Xiuping Liu \and
Yi Zhang \and
Huasheng Wang \and
Jiang Liu \and
Ying Chen \and
Hantao Liu
}
\begin{document}
\maketitle
\begin{abstract}
In this paper, we leverage the advantages of event cameras to resist harsh lighting conditions, reduce background interference, achieve high time resolution, and protect facial information to study the long-sequence event-based person re-identification (Re-ID) task.
To this end, we propose a simple and efficient long-sequence event Re-ID model, namely the Spike-guided Spatiotemporal Semantic Coupling and Expansion Network (S$^3$CE-Net). To better handle asynchronous event data, we build S$^3$CE-Net based on spiking neural networks (SNNs).
The S$^3$CE-Net incorporates the Spike-guided Spatial-temporal Attention Mechanism (SSAM) and the Spatiotemporal Feature Sampling Strategy (STFS).
The SSAM is designed to carry out semantic interaction and association in both spatial and temporal dimensions, leveraging the capabilities of SNNs.
The STFS involves sampling spatial feature subsequences and temporal feature subsequences from the spatiotemporal dimensions, driving the Re-ID model to perceive broader and more robust effective semantics. 
Notably, the STFS introduces no additional parameters and is only utilized during the training stage.
Therefore, S$^3$CE-Net is a low-parameter and high-efficiency model for long-sequence event-based person Re-ID.
Extensive experiments have verified that our S$^3$CE-Net achieves outstanding performance on many mainstream long-sequence event-based person Re-ID datasets.
\textbf{Code is available at:} \href{https://github.com/Mhsunshine/SC3E_Net}{\texttt{https://github.com/Mhsunshine/SC3E\_Net}}.
\end{abstract}    
\section{Introduction}
\label{sec:intro}

\begin{figure}[tb]
\centering
\includegraphics[scale=0.9]{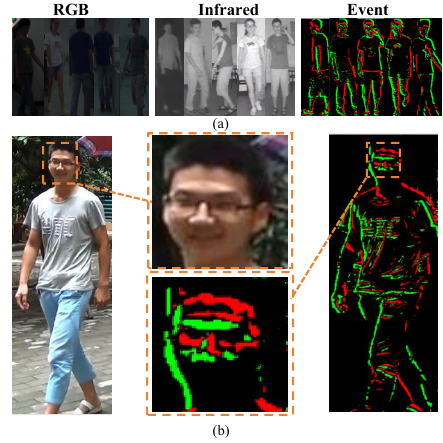}
\caption{(a) Display of person information in RGB, Infrared, and Event modalities under low-light conditions. (b) Display of facial detail comparison in RGB and Event modalities.}
\label{fig:1}
\end{figure}

Video-based person re-identification (Re-ID) aims to retrieve a specific person among numerous input video sequences \cite{9099320}.
Current methods \cite{10.1007/978-3-030-58536-5_14, 9412371, 9710753, 8675957, 10056930, 10136698} primarily rely on RGB data and design various spatiotemporal feature extraction modules to effectively address challenges in Re-ID tasks, such as viewpoint changes, occlusion, and pose variations.
However, traditional RGB cameras struggle to perceive effective person semantics under conditions of high exposure or low lighting.
Some excellent work has studied Re-ID under infrared modality \cite{DBLP:journals/pami/YeWCD24, DBLP:journals/pr/YangLWG25}.
Although these methods have achieved excellent results, both types of cameras face a common challenge: persons are easily distracted by background information, motion blur, and the exposure of their facial information.

Recently, event cameras \cite{9138762} have emerged as a promising sensor technology, capable of recording pixel-level brightness changes.
As shown in Fig.\ref{fig:1}-(a), compared to RGB/Infrared cameras, event cameras exhibit not only resistance to extreme lighting conditions but also significantly reduce interference from background information \cite{10.1007/978-3-030-58601-0_10, 10.1007/978-3-030-58598-3_41}. Furthermore, as illustrated in Fig.\ref{fig:1}-(b), event cameras offer an advantage over RGB cameras by minimizing the leakage of personal portrait information. Additionally, the high temporal resolution of event cameras ensures that motion blur is not easily induced in the event modality.
Given these advantages of the event modality, there is considerable potential for developing high-quality Re-ID models.

Despite these advantages, the event modality also presents unique challenges.
First, the spatial sparsity of event streams necessitates the association of events across different time points in order to obtain meaningful spatial features.
Additionally, the high temporal resolution of event cameras generates a large volume of event data, necessitating the development of efficient processing methods to extract useful information while avoiding excessive computational complexity.

\begin{figure}[tb]
\centering
\includegraphics[height=5cm]{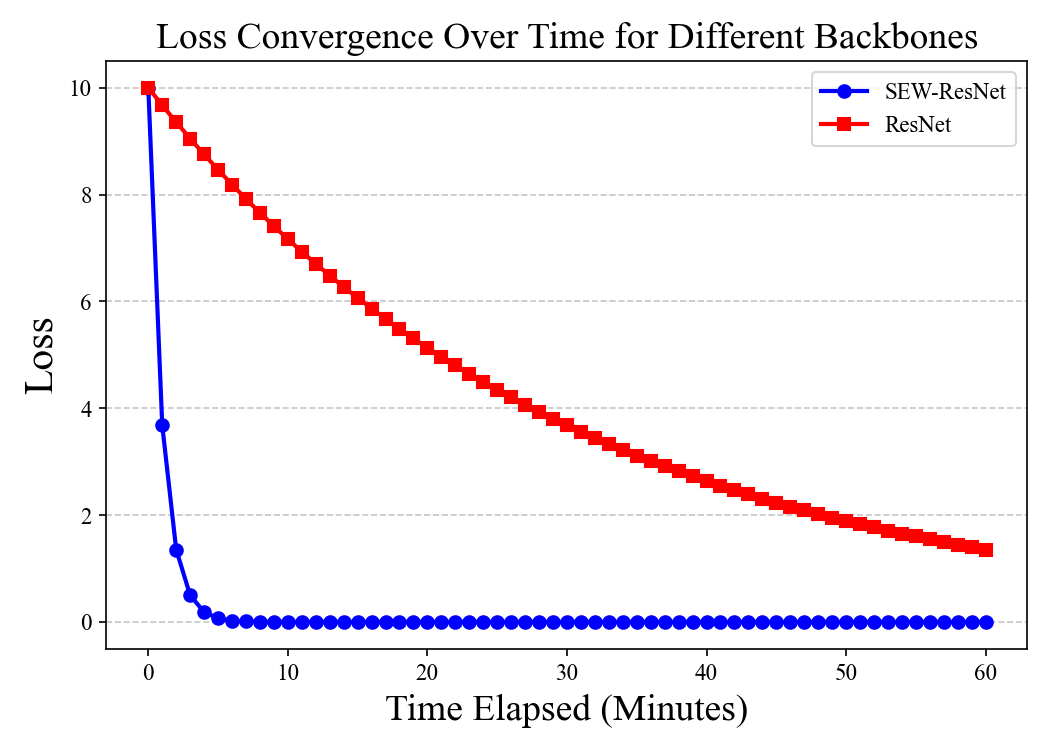}
\caption{Convergence of the SNNs' Backbone (depicted by the blue line, SEW-ResNet) and the CNNs' Backbone (depicted by the red line, ResNet). }
\label{fig:5}
\end{figure}

Current mainstream video-based Re-ID methods are predominantly based on Deep Convolutional Networks (DCNs) \cite{10.1007/978-3-030-58536-5_14, 9412371, 9710753, 8675957}.
However, DCNs face significant limitations when handling event camera data, particularly in efficiently processing sparse and asynchronous event streams. Furthermore, DCNs are often associated with high energy consumption and latency, rendering them unsuitable for resource-constrained or real-time applications.
Recent Transformer-based approaches \cite{DBLP:conf/cvpr/000100LS23, DBLP:conf/cvpr/MenapaceSSDCKFS24} attempt to achieve simultaneous interaction of temporal and spatial semantics, but they require significant computational overhead.
In contrast, Spiking Neural Networks (SNNs) \cite{MAASS19971659}, inspired by biological neural networks, transmit information through discrete spikes and only perform computations when input events are received. This design enables SNNs to achieve low latency and low power consumption, making them especially well-suited for asynchronous event streams.
Accordingly, the asynchronous nature of SNNs aligns perfectly with the characteristics of event streams \cite{PMID:25104385}, allowing them to effectively process sparse and high temporal resolution inputs while activating and accumulating temporal semantics. As shown  in Fig.~\ref{fig:5},  in the Re-ID task based on event data, SNNs' Backbone (depicted by the blue line, SEW-ResNet) converges faster than CNNs' Backbone (depicted by the red line, ResNet).
However, SNNs also face difficulties in achieving spatiotemporal correlation interactions between discrete and sparse semantics, which is not conducive to  representation of compact person descriptors.

Besides, we know that current deep models often concentrate on a small subset of key semantics while overlooking other valid semantics. For event modalities with extremely sparse semantics, the "lazy" nature of deep models can easily render person descriptors less robust.
A series of neural models adopt the feature dropout strategy \cite{Nitish2014, Jonathan2015, Golnaz2018, Zuozhuo19} in an attempt to improve the robustness of the model. Additionally, BCP \cite{Yifan2018} extracts strip-shaped patch features to perceive and capture effective semantics.
However, due to the extreme sparsity of event semantics, the dropout strategy often falls into empty areas with a high probability, and the semantic information contained within the strip-shaped regions may be too limited to meet the recognition requirements.


To solve the above two issues, we propose a simple and efficient long-sequence event Re-ID model, the Spike-guided Spatiotemporal Semantic Coupling and Expansion Network (S$^3$CE-Net).
To better perceive and accumulate key semantics efficiently from sparse and asynchronous event streams, we adopt the Spike element-wise ResNet (SEW-ResNet) \cite{Fang2021DeepRL} to build our S$^3$CE-Net. 
The S$^3$CE-Net contains the Spike-guided Spatiotemporal Attention Mechanism (SSAM) and Spatiotemporal Feature Sampling Strategy (STFS). In SSAM, we first use SNNs to perceive and accumulate temporal semantics. Then, with the help of historical temporal semantics, the semantics of the current moment are interacted and fused across time and space. Note that, in the time dimension, the event information of future moments is unknown. Therefore, in the process of enhancing the semantic meaning of the current moment, we do not consider the future spatial and temporal semantics. This design facilitates real-time computation in practical scenarios.
To better leverage the advantages of SSAM, we introduce it at different levels of SEW-ResNet. With the help of SSAM and SEW-ResNet, isolated and discrete spatiotemporal semantics would be better correlated.
In STFS, we randomly sample a number of large-sized patches in the spatial dimension to cover the original features and drive the model to perceive and capture effective semantics from each patch. Accordingly, we randomly collect event sub-sequences from the time dimension. The combination of these two strategies further enhances the model's ability to capture robust semantics. STFS has no additional parameters and is only used during the training stage.
Therefore, S$^3$CE-Net is a low-parameter and high-efficiency long-sequence event person Re-ID model.
The main contributions of this paper are as follows:
\begin{itemize}
    \item We proposed the   SSAM  to enhance the interaction and dependence of spatiotemporal semantics. 
    
    \item We  introduce the STFS  to  drive  the Re-ID model to capture broader and more robust spatiotemporal semantics.
    
    \item  Combined SSAM and STFS,  we propose   a  simple  and efficient  long-sequence event  ReID network, S$^3$CE-Net.   Our S$^3$CE-Net achieves  outstanding performance on many  mainstream long-sequence even  ReID datasets. 
\end{itemize}

\begin{figure*}[tb]
	\centering
	\includegraphics[height=11.5cm]{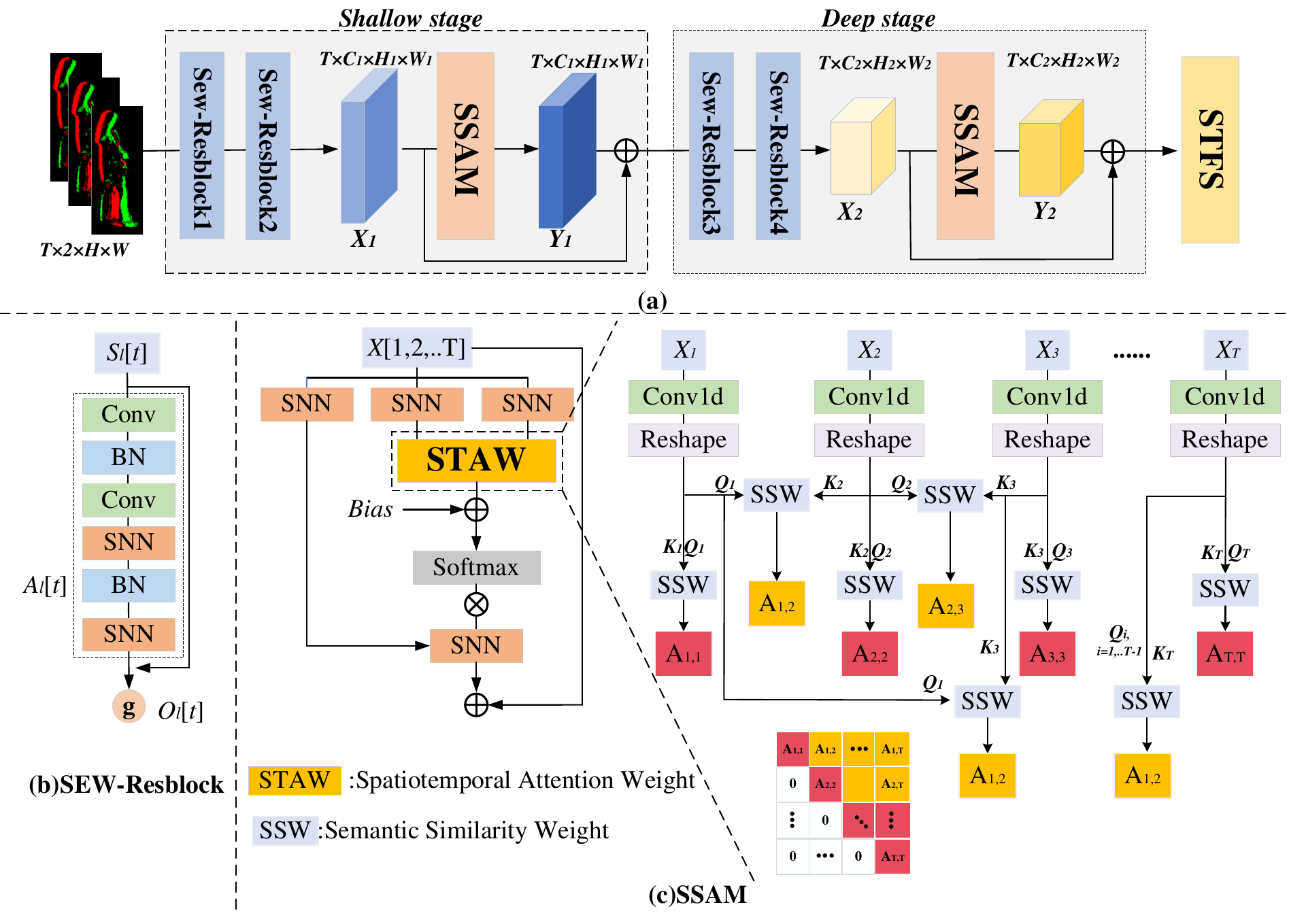}
	\caption{Overview of our S$^3$CE-Net (a) which takes $T$ event tensors as input, using SEW-ResNet18 as the backbone. It extracts features through four Spike-Element-Wise Residual blocks (SEW-Resblock) (b) and establishes spatiotemporal  dependencies via the Spike-guided Spatiotemporal Attention Mechanism (SSAM) (c). Finally, the Spatiotemporal Feature Sampling Strategy (STFS) module drives model perceives broader and more robust effective semantics.}
	\label{fig:2}
\end{figure*}
\section{Related Work}
\label{sec:formatting}

\subsection{Video-Based Person Re-Identification}
Video-based Person Re-identification (Re-ID)\cite{10.1007/978-3-030-58595-2_24,10.1007/978-3-030-58568-6_20,9156659,9416694} extends traditional Image-based Re-ID methods\cite{10054607,10049924,9874750,9710089,9578119,9710001,9578051,9009039} by leveraging the temporal information in videos to improve person matching accuracy. 
Compared to single-frame images, videos contain rich and useful information, including temporal and spatial features, as well as variations in body posture from different viewpoints.

\textit{Global appearance-based methods} often aggregate multi-frame features into a single global feature vector using pooling\cite{10.1007/978-3-319-46466-4_52,8578144,2018Revisiting} or recurrent neural networks (RNNs)\cite{7780517,Liu_Yuan_Zhou_Li_2019}.  
\textit{To better preserve temporal information}, spatiotemporal attention mechanisms and graph convolutional networks (GCNs) have been introduced to enhance matching accuracy by focusing on key regions or establishing relationships between frames. 
Spatiotemporal attention mechanisms, such as the STA
\cite{Fu_Wang_Wei_Huang_2019}, focus on the most discriminative parts of the video, mitigating the effects of lighting changes and occlusion on recognition. 
Graph convolutional networks (e.g., SGGNN\cite{8579000}) leverage graph structures to learn the similarity relationships between frames, enhancing the model's robustness.
Some studies also propose \textit{local part alignment methods}, which extract features from local regions in the video to avoid misalignment between frames. 
For instance, spatiotemporal completion networks (STCnet)\cite{8954276} tackle partial occlusion by recovering the appearance information of occluded parts. 
To extract fine-grained cues, \cite{9156850} proposed Multi-Granularity Reference Assisted Feature Aggregation (MG-RAFA) to jointly process spatiotemporal features.
Meanwhile,  STT \cite{Zhang2021SpatiotemporalTF}  adopted  the \textit{Transformer pattern}   to  enhance the representation of  long-range temporal dependencies  by processing relationships between frames.





\subsection{Event-based Vision Tasks}
As new types of  sensors,  in  high-speed motion, event cameras can capture brightness changes with microsecond-level response times, effectively reducing motion blur\cite{10377805,10205073,9710726,10.1007/978-3-031-19797-0_24}. 
Their asynchronous data acquisition allows for lower latency and faster response in tasks such as Visual Odometry (VO)\cite{Zuo2022DEVODC,Mueggler2017ContinuousTimeVO,Zhu2017EventBasedVI,Klenk2023DeepEV} and SLAM\cite{Weikersdorfer2013SimultaneousLA,Chamorro2022EventBasedLS}. Besides, event cameras maintain excellent visual perception   under    high or low-light conditions.    
This enables event cameras to perform excellently in tasks such as optical flow estimation\cite{Gehrig2021ERAFTDO,Lee2020SpikeFlowNetEO,Ding2021SpatioTemporalRN}, object detection\cite{8593805}, and tracking\cite{Gehrig2018EKLTAP,PMID:32848527}, helping achieve stable and reliable visual tasks under complex lighting conditions.
For Re-ID, event cameras can reduce the leakage of person appearance information, and the imaging principle of event cameras can reduce the interference of background information.

\section{Methods}

\subsection{Overview}
Our  proposed  Spike-guided Spatiotemporal Semantic Coupling and Expansion Network (S$^3$CE-Net) is illustrated in Fig.~\ref{fig:2}. 
We adopt the SEW-ResNet\cite{Fang2021DeepRL} as the backbone of our S$^3$CE-Net.
The event tensor sequence  is the  input information for S$^3$CE-Net. 
The  proposed S$^3$CE-Net contains the  Spike-guided  Spatialtemporal Attention Mechanism (SSAM) and Spatiotemporal Feature Sampling Strategy (STFS). 
The SSAM is designed to  conduct  semantic interaction and association in the spatial and  temporal dimensions. To better leverage the advantages of SSAM, we integrate SSAM at different levels of SEW-ResNet.   
In the STFS,  we sample patches in the spatial dimension and sub-sequences in the temporal dimension, driving the model to perceive the features of each region equally to capture robust semantics.
The  STFS is parameter-free and is  applied only during the training stage to improve model robustness;   SSAM is also built on SNNs.   The computation of parameters only occurs in SSAM and the backbone SEW-ResNet. So, S$^3$CE-Net can be considered a lightweight network.

\subsection{Event Representation}
Event stream captured by the event camera is represented by a set \(  E=\left\{e_{i} \mid i=0,1,2, \ldots, N\right\}\), with a single event denoted as \(e_{i}=(x_{i},y_{i},t_{i},p_{i})\), where \( x_{i}\), \(y_{i}\) represent the spatial coordinates, \(t_{i} \) is the timestamp, and \(p_{i}\) indicates the direction of the brightness change (increase  or decrease).
To this, we divide the asynchronous event stream into a sequence of \(T\) tensors with fixed time intervals, where each tensor consists of two channels: one for positive polarity events and one for negative polarity events. So, the processed  event tensors  is  $X_e \in R^{T \times 2 \times W  \times H}$.

\subsection{Backbone: SEW-ResNet}
Spiking Neural Networks (SNNs) excel in processing event-driven data, such as the sparse and asynchronous signals generated by event cameras. 
Furthermore,  SEW-ResNet\cite{Fang2021DeepRL} leveraged the Spike-Element-Wise Residual block (SEW-Resblock) to efficiently implement residual learning and stabilize gradient propagation. For this reason,  we adopt SEW-ResNet as  our  S$^3$CE-Net's backbone.  
Structure of SEW-Resblock  is shown in  Fig.~\ref{fig:2}-b. For more details about SEW-ResNet,  refer to \cite{Fang2021DeepRL}.

\subsection{SSAM}
As previously mentioned in Intro.~\ref{sec:intro},  compared with CNNs and Transformers,  SNNs are good at processing time-series event streams.
However, event semantics are extremely discrete and isolated. 
SNNs are difficult to effectively model the association and dependence between discrete semantics.
The compact contextual semantics is more conducive to driving the model to efficiently perceive and capture discriminative person semantics.
Thus,  based on  SNNs,  we  further conduct  spatiotemporal semantic association modeling  to enhance the compactness between semantics.  However, in the time dimension, the event information of future moments is unknown. Therefore, we only use historical temporal semantics to interact and integrate the semantic of events at the current moment.   It differs from the Transformer, which performs semantic interaction across all time steps.
This is more aligned with practical real-time application scenarios.
To achieve the goal, we propose a Spike-guided Spatiotemporal Attention Mechanism (SSAM).

As shown in  Fig.~\ref{fig:2},  to better leverage the advantages of SSAM, we introduce SSAM at two stages of SEW-ResNet.   
For the input feature \(X_s= (x_{s,1}, x_{s,2}, \cdots, x_{s,T}) \in R^{T\times C_s\times H_s\times W_s} \), \( s\in \{1,2\}\), where  $x_{s,t}$ represents the event semantics corresponding to the $t$-th time step, \(s\) represents the \(s\)-th feature extraction stage of SEW-ResNet.  We first flatten feature $X_s$ by  $X_s=Flatten(X_s)\in R^{T\times C_s\times(H_s\times W_s)}$.

Then, we  calculate the Spatiotemporal Attention Weight (STAW) (as shown in Fig.~\ref{fig:2}-c) by 
\begin{equation}
Q_{s,global}= SNN_{Q}(Conv_{1d}(X_s))
\end{equation}
\begin{equation}
K_{s,global}= SNN_{K}(Conv_{1d}(X_s))
\end{equation}
\begin{equation}
V_{s,global}= SNN_{V}(Conv_{1d}(X_s))
\end{equation}
Where $SNN_{Q}$, $ SNN_{K}$, and $SNN_{V}$ are separate SNN Layers  with independent parameters, $Conv_{1d}$  is the $1\times 1$ convolution operation,  \( Q_{s,global}, K_{s,global}, \text{and } V_{s,global} \in R^{T\times C_s\times N_s},  N_s = H_s \times W_s \) are the global query, key, and value obtained through the \(s\)-th stage, representing the features across the entire temporal sequence. 
From $Q_{s,global}$, $K_{s,global}$, and $V_{s,global}$, we extract the query, key, and value for each tensor $t$-th time step as: 
$Q_{s,t}=Q_{s,global}[t, :, :], K_{s,t}=K_{s,global}[t, :, :], \text{and}   V_{s,t}=V_{s,global}[t, :, :]$, 
where $Q_{s, t}$, $K_{s, t}$, $V_{s,t} \in R^{C_s \times N_s}$ represent the query, key, and value for event tensor $t$, $t \in \{1, 2, \dots, T\}$ is the time step index, $N_s =H_s \times W_s$ represents the number of tokens  per  event tensor.

To capture the spatial relationships  among $N_s$ tokens within an event tensor $x_{s,t}$, we introduce a learnable intra-frame relative position bias $B \in R^{N_s\times N_s} $, which captures the spatial dependencies within a single event tensor. A detailed formulation of $B$ and its implementation is provided in Section $A$  of   Supplementary Materials.



When computing the attention matrix, each event tensor at a given time step interacts only with  tensors from the current and previous time steps. 
This is because the value weights of future tensor information with respect to current information are unknown, ensuring temporal consistency in the model and aligning with the real-world understanding of dynamic scenes.  
This is different from the previous Transformer idea and will effectively reduce the amount of computation.
For the convenience of expression, we omit the variable $s$ in the tensor calculation stage without causing ambiguity.  
So,  the similarity weight of the event tensor at the $t'$-th time step and the event tensor at the $t$-th time step (current time step) is calculated as:
\begin{equation}
A_{t',t} = 
\begin{cases} 
\frac{Q_{t'} \cdot K_t^T}{\sqrt{D}}  & \text{if } t' < t \\
 \frac{Q_{t'} \cdot K_t^T}{\sqrt{D}} + B& \text{if } t' = t\\ 0&\text{if } t' > t
\end{cases}
\end{equation}
Here, for the current time step $t$, when \(t'\leq t\), we retain the calculated weight, otherwise, the weight is reset to $0$ to discard the influence of future event tensors on the current moment.
Since this, the STAW shown in Fig.~\ref{fig:2}-(c) is the final similarity weight matrix we calculate. We can clearly see that the weight matrix is an upper triangular matrix.

Finally, based on such a weight matrix, we perform interaction and fusion between spatiotemporal semantics by 
\begin{equation}
O_t =  SNN(\sum_{t'=1}^{t}Softmax(A_{t', t},dim= -1) \cdot V_t)
\end{equation}
where $O_t \in R^{C_s \times N_s}$ represents the output that  fuses the past spatiotemporal semantic information related to the $t$-th event tensor.  The final output is obtained by concatenating the outputs of all  event tensors:
\begin{equation}
Y_s=[O_1; O_2; \cdots; O_T] \in R^{T\times C_s\times N_s}
\end{equation}




\begin{figure}[t!]
\centering
\includegraphics[scale=0.7]{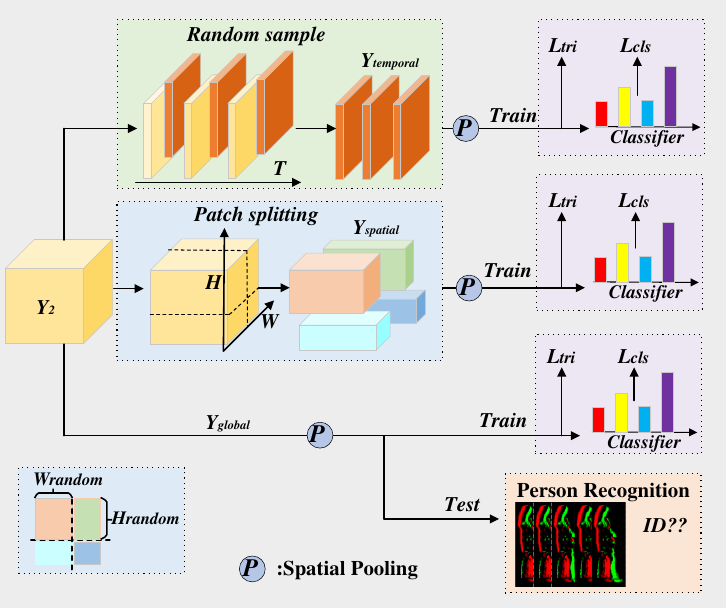}
\caption{Architecture of the proposed STFS. For different branch, we use triplet loss and label smoothing cross-entropy loss for constraints, represented by $L_{tri}$ and $L_{cls}$.}
\label{fig:3}
\end{figure}

\subsection{STFS}
As mentioned in Intro.~\ref{sec:intro}, deep models can easily fall into local effective semantics, ignoring  perception of broader effective semantics. To this, we try to use sub-regions and sub-sequences from the spatial and temporal dimensions to drive deep models to perceive broader effective semantics. 
So,  we propose the Spatiotemporal Feature Sampling Strategy (STFS).  Schematic diagram of STFS is shown in Fig.~\ref{fig:3}. 

To reduce the computational load and effectively feed back the training of the entire network, our STFS is applied on the deep feature map $Y_2$.
As  shown  in Fig.~\ref{fig:3},  from top to bottom, they are subsequence sampling in the time dimension, subregion sampling in the spatial dimension, and the extraction branch of global semantic descriptors.

\hspace{1em}\textbf{Subsequence Sampling in the Time Dimension: }
As  shown in top branch of  Fig.~\ref{fig:3}, we randomly select $k = \frac{T}{2}$ time steps \(T_{random}\subset\{1, 2, \cdots, T\}\), which can be represented as:
\begin{equation}
T_{random}=\{t_{1},t_{2},\cdots,  t_{k}\}
\end{equation}
Perform average pooling on each selected time step to obtain the feature representation corresponding to each random time step.
\begin{equation} 
\begin{aligned}
   &Y_{temporal} = \{ V_{t_{i}} \mid t_{i} \in T_{random} \}\\ 
   &V_{t_{i}}=Avg2d\left(Y_2[t_{i}, :, :, :]\right)
\end{aligned}
\end{equation}
where $Y_{temporal}$ represents the randomly selected subsequence samples' semantics in the time dimension,  \(Avg2d\) refers to the 2D average pooling operation.

\hspace{1em}\textbf{Subregion Sampling in the Spatial Dimension:}
The feature map is divided into four random patches, and average pooling is performed separately on each patch. 
The average pooling operation in each patch results in a feature vector that represents the features of the local region.
\begin{equation} 
\begin{aligned}
&Y_{spatial} = \{ V_{ul}, V_{ur}, V_{ll}, V_{lr} \}, \\
&V_{ul}=Avg2d(Y_{2}[:, :, :W_{random}, :H_{random}]).  \\
\end{aligned}
\end{equation}
where \(ul, ur, ll, lr\) denote the upper left, upper right, lower left, and lower right feature regions respectively; \(W_{random}\) and \(H_{random}\) are obtained by random sampling on \([0,W_2],[0,H_2] \) respectively.
Similarly, the other regions—upper right (\(V_{ur}\)), lower left (\(V_{ll}\)), and lower right (\(V_{lr}\))—are obtained by performing average pooling on the corresponding patches of the feature map.

\hspace{1em}\textbf{Global Semantic Extraction:}
At the 2-th stage, we restore the feature $Y_2$ to its spatiotemporal dimensions, and  obtain the global semantic vector by 
\begin{equation}
\begin{aligned}
  &Y_{2}=Reshape(Y_2)\in R^{T\times C_2\times H_2\times W_2},\\  
  &Y_{global}=Avg2d(Y_{2}[:, :, :, :]). 
\end{aligned}
\end{equation}
During the training phase, we use triplet loss $L_{tri}$  and cross-entropy loss with label smoothing $L_{cls}$  to constrain $Y_{temporal}$, $Y_{spatial}$, and $Y_{global}$.The  loss function of our $S^3$CE-Net is summarized as 
$L_{total} = \lambda_1 L_{tri} + \lambda_2 L_{cls}$.  The $\lambda_1=1.0$ and $\lambda_2=0.1$ represent the weights of different losses. 
During the testing stage,  we only  use  the  global semantic  vector  $Y_{global}$  to conduct  the person matching.



\section{Experiments}

\begin{table}\small
  \caption{Performance (\%) comparison with various SOTA methods on MARS, PRID-2011, and iLIDS-VID datasets.}
  \label{tab:comparison1}
  \centering
  \renewcommand{\arraystretch}{1}  
  \setlength{\tabcolsep}{0.5mm}  
  \begin{tabular}{cc|cc|cc|cc}
    \toprule[1pt]
    \multicolumn{2}{c|}{\textbf{Methods}} & \multicolumn{2}{c|}{\textbf{MARS}} & \multicolumn{2}{c|}{\textbf{PRID-2011}} & \multicolumn{2}{c}{\textbf{iLIDS-VID}} \\

    \cline{1-8} 
     Network & Input& mAP & Rank-1 & mAP & Rank-1 & mAP & Rank-1 \\ 
    \hline
    GRL\cite{Liu2021WatchingYG} & E & 24.7 & 16.7 & 22.3 & 11.8 & 25.2 & 13.2   \\ 
    OSNet\cite{Zhou2019OmniScaleFL}& E & 20.9 & 19.3 & 22.2 & 10.1 & 23.7 & 12.6  \\
    SRS-Net\cite{Wang2020SimpleAE}& E & 17.9 & 10.6 & 19.2 & 11.3 & 20.7 & 9.3 \\
    STMN\cite{Eom2021VideobasedPR}& E & 22.4 & 10.9 & 20.2 & 11.2 & 21.1 & 10.4 \\
    CTL\cite{Liu2021SpatialTemporalCA}& E & 19.6 & 12.7 & 20.4 & 12.4 & 24.1 & 12.5\\
    PSTA\cite{Wang2021PyramidSA}& E & 22.7 & 12.0 & 21.4 & 12.5 & 22.4 & 10.0\\ 
    FastReID\cite{He2020FastReIDAP} & E & 39.3 & 26.2 & 65.8 & 54.1 & 41.8 & 29.6\\
    SDCL \cite{10203615} & E & 38.3 & 26.8 & 61.1 & 50.7 & 43.1 & 30.7\\
    VSLA-CLIP \cite{Zhang2024CrossPlatformVP} & E & 33.4 & 20.9 & 57.5 & 45.3 & 36.9 & 26.6\\
    TF-CLIP\cite{Yu2023TFCLIPLT} & E & 41.9 & 29.7 & 64.7 & 52.4 & 44.2 & 32.6\\
    CMTC\cite{Li2025EventbasedVP} & E & 43.8 & 31.6 & 71.9 & 60.4 & 47.5 & 35.3 \\
    \hline
    Ours& E & \textbf{67.4} & \textbf{55.2} & \textbf{90.3} & \textbf{80.9} & \textbf{71.0} & \textbf{58.8}\\ 
    \bottomrule[1pt] 
  \end{tabular}
\end{table}

\subsection{Datasets}
Currently, there is no event-based long-sequence ReID dataset. 
We first follow Cao's\cite{10203615} to adopt the Video-to-Event \cite{9523069} method for generating event streams from three classic video-based person Re-ID datasets,  \textbf{PRID-2011}\cite{10.5555/2009594.2009606}, \textbf{iLIDS-VID}\cite{10.1007/978-3-319-10593-2_45}, and \textbf{MARS}\cite{10.1007/978-3-319-46466-4_52}. 
In addition, we conduct experiments on the first \emph{real} event-camera dataset for Re-ID, namely \textbf{Event-ReID}\cite{10377578}. 
Event-ReID was captured by four indoor 640\,$\times$\,480 Prophesee cameras and contains 33 identities walking across the four views. 
After hand-crafted and YOLO-assisted annotation, we obtain 16\,k bounding boxes that form long event streams suitable for sequence-level evaluation. 
Comprehensive statistics, annotation protocol and preprocessing details are provided in the Supplementary Materials.

\begin{table}[t]
  \setlength{\belowcaptionskip}{1pt}
  \centering
  \small        
  \caption{Performance (\%) on the \textit{Event-ReID} dataset.}
  \label{tab:event_reid_all}
  \renewcommand{\arraystretch}{0.9}     
  \setlength{\tabcolsep}{3pt}           

  \begin{tabular}{@{}lSS@{}}
    \toprule
    \textbf{Method} & \textbf{mAP} & \textbf{Rank-1} \\
    \midrule
    ResNet-18  & 71.2 & 60.5 \\
    \textbf{Ours} & \textbf{93.7} & \textbf{86.3} \\
    \bottomrule
  \end{tabular}
\end{table}



\subsection{Implementation Details}
Our network is implemented based on the PyTorch deep learning framework on a single NVIDIA RTX 2080Ti GPU. The training process lasts for 100 epochs, with an initial learning rate of 0.00035, which is decayed to one-third of its original value every 30 epochs. The Adam optimizer is used for parameter updates. During the testing phase, cosine similarity is employed to measure the distance between gallery and query instances.

\begin{figure}[tb]
\centering
\includegraphics[height=5.5cm]{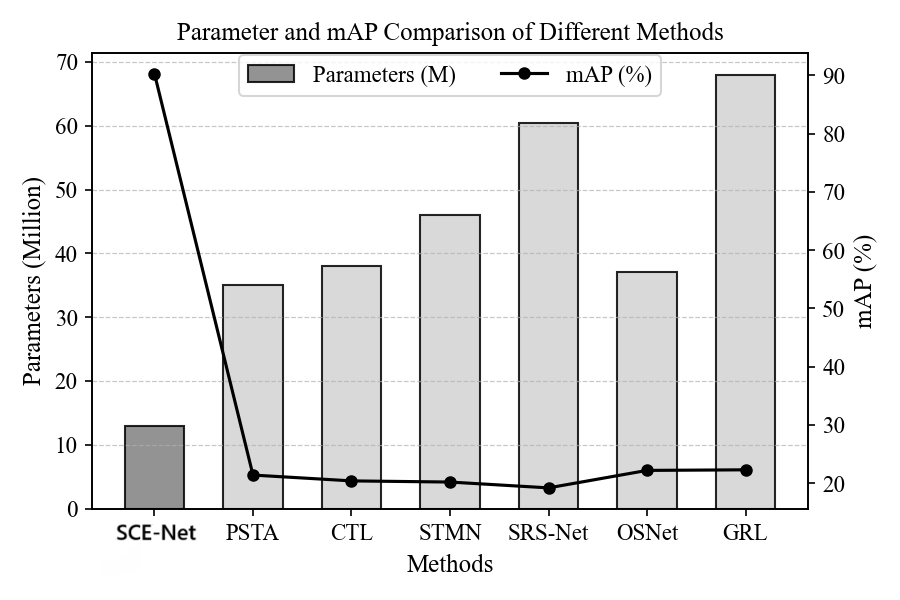}
\caption{Parameters and mAP (in \%) comparison.}
\label{fig:4}
\end{figure}

\subsection{Comparison with SOTA methods}
Table~\ref{tab:comparison1} shows the performance of our proposed method for event-driven video person re-identification on MARS, PRID-2011, and iLIDS-VID datasets. 
Due to the scarcity of existing event-driven video person re-identification methods, we will retrain and test the previous RGB-based Re-ID model with event data.
As shown in Table~\ref{tab:comparison1},  our network achieves excellent performance across multiple public datasets, further demonstrating its advanced nature and effectiveness on event data.
The experimental results indicate that current methods developed for RGB data do not perform well with asynchronous event data. 
Besides, Fig.~\ref{fig:4} clearly demonstrates that our S$^3$CE-Net achieves higher accuracy  with fewer parameters compared to other methods.

To further demonstrate the robustness of the model under blurred conditions, we followed Cao's\cite{10203615} method to introduce blur into the PRID-2011 and iLIDS-VID datasets to simulate scenarios with fast motion. 
As shown in Table~\ref{tab:comparison2}, for a fair comparison, the experiments for other SOTA methods were divided into two groups: the first group used the original RGB Video  data as input (i.e. V), and the second group used both RGB and Event data as input (i.e. V+E). 
The experimental results in Table~\ref{tab:comparison2} clearly demonstrate the advantages of event data under degraded conditions. 
On the PRID dataset, our method achieved the best performance, highlighting the rationality of our model framework designed specifically for event data. 
Limited by the fact that event data is more sparse and has lower information density compared to RGB data, coupled with the frequent person  occlusions in the iLIDS-VID dataset, the extraction of target features becomes constrained, increasing the difficulty of feature extraction and recognition. 
Consequently, compared to the SDCL\cite{10203615}, we achieved comparable performance levels using only event data. This means that our model can mine effective person  semantics from event streams for person matching.

\begin{table}\small
  \caption{The mAP(\%) values of different methods on PRID-2011, and iLIDS-VID datasets in blurry conditions.
}
  \label{tab:comparison2}
  \centering
  \renewcommand{\arraystretch}{1}  
  \setlength{\tabcolsep}{3mm}  
  \begin{tabular}{cc|c|c}
    \toprule[1pt]
    \multicolumn{2}{c|}{\textbf{Methods}}  & \multicolumn{1}{c|}{\textbf{PRID-2011}} & \multicolumn{1}{c}{\textbf{iLIDS-VID}} \\

     \cline{1-4} 
     Network & Input& Blurry  & Blurry  \\ 
    \hline
    GRL\cite{Liu2021WatchingYG} & V  & 81.8 & 60.7  \\ 
    SRS-Net\cite{Wang2020SimpleAE}& V  & 78.8 & 61.6  \\ 
    STMN\cite{Eom2021VideobasedPR}& V  & 81.5 & 57.7  \\ 
    CTL\cite{Liu2021SpatialTemporalCA}& V  & 81.6 & 56.0  \\ 
    PSTA\cite{Wang2021PyramidSA}& V  & 79.7 & 58.9  \\ 
    \hline
    GRL\cite{Liu2021WatchingYG} & V+E & 88.4 & 66.2  \\ 
    SRS-Net\cite{Wang2020SimpleAE}& V+E  & 82.1 & 67.8  \\ 
    STMN\cite{Eom2021VideobasedPR}& V+E  & 85.5 & 62.3  \\ 
    CTL\cite{Liu2021SpatialTemporalCA}& V+E  & 88.6 & 64.2  \\
    PSTA\cite{Wang2021PyramidSA}& V+E  & 85.0 & 65.5  \\ 
    SDCL\cite{10203615}& V+E  & 89.5 & \textbf{71.4} \\
    \hline
    Ours& E & \textbf{90.1} & 70.4 \\ 
    \bottomrule[1pt] 
  \end{tabular}
\end{table}
\subsection{Experimental on the Real Event-ReID Dataset}
On the real Event-ReID dataset (Table \ref{tab:event_reid_all}), our method surpasses the ResNet-18 baseline by 22.5 pp in mAP and 25.8 pp in Rank-1, confirming the efficacy of SSAM and STFS on sparse event streams. SSAM extracts coherent spatiotemporal semantics and suppresses noise, STFS randomly samples diverse sub-blocks to mitigate over-fitting, and the event-driven SNN naturally matches asynchronous data. As Event-ReID contains only 33 identities, future work will validate generalisation on larger, cross-scene datasets.

\subsection{Ablation Study}
\subsubsection{Effectiveness of SNN:}
In our S$^3$CE-Net, we employed a SNN to extract person  feature from  event streams. 
To validate the rationality and effectiveness of introducing SNNs, we performed ablation experiments by replacing the SNN model with a classic ResNet. 
The experimental results in Table~\ref{tab:comparison3} indicate that as the network depth increases, the performance of the model is gradually declining. 
This suggests that SNNs  possess a superior capability for capturing and extracting event semantics in comparison to CNNs.

\begin{table}\small
  \caption{Impact of different Baselines on  Re-ID performance.}
  \label{tab:comparison3}
  \centering
  \renewcommand{\arraystretch}{1}  
  \setlength{\tabcolsep}{5.4mm}  
  \begin{tabular}{c|c|c}
    \toprule[1pt]
    \multicolumn{1}{c|}{\textbf{Network}}  & \multicolumn{1}{c|}{\textbf{PRID-2011}} & \multicolumn{1}{c}{\textbf{iLIDS-VID}} \\

     \cline{1-3} 
    
    \hline
    Sew-Resnet  & 61.9 & 50.2  \\ 
    Resnet-18  & 35.5 & 31.2  \\ 
    Resnet-50  & 32.0 & 30.3  \\ 

    \bottomrule[1pt] 
  \end{tabular}
\end{table}

\subsubsection{Effectiveness of SSAM and STFS:}
The  SSAM is  designed   to enhance the interaction and dependence of spatiotemporal semantics.  
The  STFS is  designed  to  drive  model perceives broader and more robust effective semantics.
The  results are  reported in Table~\ref{tab:ablation1}. 
As  shown in Table~\ref{tab:ablation1},  we can clearly see that SSAM and STFS have significant contributions to the performance of S$^3$CE-Net.  The numerical experimental results significantly verify the effectiveness of the two designs.

\begin{table}[]\small
	\caption{Results produced by combining different mechanisms in the S$^3$CE-Net. }
	\label{tab:ablation1}
	\centering
	\renewcommand{\arraystretch}{1}
	\setlength{\tabcolsep}{3.2mm}{\begin{tabular}{c|c |c c |c c}
			\toprule[1pt]
			 \multirow{2}{*}{SSAM} & \multirow{2}{*}{STFS}  & \multicolumn{2}{|c|}{\textbf{PRID-2011}} & \multicolumn{2}{|c}{\textbf{iLIDS-VID}}\\ 
			\cline{3-6} 	
			  &   & mAP & rank-1 & map & rank-1  \\ 
              \hline
		   w/o & w/o    & 61.9 & 52.0 & 50.2 & 43.3  \\
			w   & w/o   & 80.2 & 73.3 & 65.6 & 57.0 \\
			w/o & w      & 67.0 & 56.1 & 51.2 & 45.5  \\ \hline
			w   & w     & \textbf{90.3} & \textbf{80.9}  & \textbf{71.0} & \textbf{58.8} \\
			\bottomrule[1pt] 
	\end{tabular}}
\end{table}

\begin{table}[]\small
	\caption{The ablation study results demonstrating the impact of incorporating SSAM at various stages.}
	\label{tab:ablation2}
	\centering
	\renewcommand{\arraystretch}{1}
	\setlength{\tabcolsep}{0.6mm}
	\resizebox{\linewidth}{!}{%
		\begin{tabular}{c|c|c|c c |c c }
			\toprule[1pt]
			\multirow{5}{*}{\textbf{SSAM}} & \multirow{2}{*}{\textbf{Shallow-stage}} & \multirow{2}{*}{\textbf{Deep-stage}}  & \multicolumn{2}{|c|}{\textbf{PRID-2011}} & \multicolumn{2}{|c}{\textbf{iLIDS-VID}}  \\ 
			\cline{4-7} 	
			& &  & mAP & rank-1 & mAP & rank-1   \\ 
            \cline{2-7}
			 & - & -  & 67.0 & 56.1 & 51.2 & 45.5  \\
			 &\checkmark & -  & 86.3 & 77.2 & 63.3 & 52.3  \\ 
			 &- & \checkmark  & 85.7 & 76.0 & 64.8 & 53.2 \\
			 &\checkmark   & \checkmark & \textbf{90.3} & \textbf{80.9} & \textbf{71.0} & \textbf{58.8}   \\ 
			\bottomrule[1pt] 
	\end{tabular}%
	}
\end{table}

\begin{table}[ht]\small
    \centering
    \caption{Results of Ablation Study on Attention Matrix in SSAM Module, STAW represents the Spatiotemporal Attention Weight shown in Fig~\ref{fig:2}-c, FAW represents the Full Attention Weight, where each time tensor computes attention with other tensors, and ZAW represents the Zero Attention Weight, where all attention weights are zero.}
    \setlength{\tabcolsep}{4.5mm}
    \label{tab:5}
    \begin{tabular}{@{}lcccccc@{}}
        \toprule
        \multirow{2}{*}{Strategy} & \multicolumn{2}{c}{PRID} & \multicolumn{2}{c}{iLIDS-VID} \\
        \cmidrule(lr){2-3} \cmidrule(lr){4-5}
                                  & MAP  & Rank-1 & MAP  & Rank-1 \\
        \midrule
        STAW   & \textbf{90.3} & \textbf{80.9}   & \textbf{71} & \textbf{58.8}   \\
        FAW    & 87.6 & 77.8   & 68.1 & 55.3   \\
        ZAW   & 67.0 & 56.1   & 51.2 & 45.5   \\
        \bottomrule
    \end{tabular}
\end{table}

			

\subsubsection{Discussion on  SSAM}
We firstly validate its effectiveness at different stages of feature extraction. 
As shown in Table~\ref{tab:ablation2} and Fig.~\ref{fig:2}, 
when the SSAM module is applied only to the shallow or deep stage, the model performance improves significantly compared to when the module is not applied. 
Specifically, for complex Re-ID datasets like iLIDS-VID, the performance improvement is more pronounced across both stages. 
This suggests that in practical applications, the stages for adding the SSAM module can be flexibly chosen based on the dataset characteristics and computational cost. 
To validate the effectiveness of the \textit{Spatiotemporal Attention Weight (STAW)} in  SSAM module, we compared it with commonly used Full Attention Weights (where each time tensor computes attention with all other  tensors, i.e. FAW in Table~\ref{tab:5}) and Zero Attention Weights (ZAW in Table~\ref{tab:5}).
The experimental results~\ref{tab:5} demonstrate that attention computation significantly improves recognition accuracy, with the design of our STAW achieving the best performance.

\begin{table}[]\small
	\caption{Ablation study results demonstrate the impact of the temporal and spatial branches in STFS.}
	\label{tab:ablation3}
	\centering
	\renewcommand{\arraystretch}{1}
	\setlength{\tabcolsep}{1.5mm}{\begin{tabular}{c|c|c|c c |c c }
			\toprule[1pt]
			\multirow{5}{*}{\textbf{STFS}} & \multirow{2}{*}{\textbf{Spatial}} & \multirow{2}{*}{\textbf{Temporal}}  & \multicolumn{2}{|c|}{\textbf{PRID-2011}} & \multicolumn{2}{|c}{\textbf{iLIDS-VID}}  \\ 
			\cline{4-7} 	
			& &  & mAP & rank-1 & mAP & rank-1   \\ 
            \cline{2-7}
			
			 &\checkmark & -  & 88.1 & 78.2 & 67.8 & 56.2  \\ 
			 &- & \checkmark  & 89.3 & 79.8 & 68.2 & 57.7 \\
			 &\checkmark   & \checkmark & \textbf{90.3} & \textbf{80.9} & \textbf{71.0} & \textbf{58.8}   \\ 
			\bottomrule[1pt] 
	\end{tabular}}
\end{table}

\subsubsection{Discussion on STFS}
We validated the contributions of its temporal and spatial branches to the model. 
Table~\ref{tab:ablation3} demonstrates that random sampling of temporal and spatial features in the final stage enhances the model's robustness without increasing its parameter size. This performance enhancement is attributed to the STFS-driven model, which expands the perception and capture range of effective semantics, thereby enhancing the robustness of semantics.

\begin{figure}[t!]
\centering
\includegraphics[height=6cm]{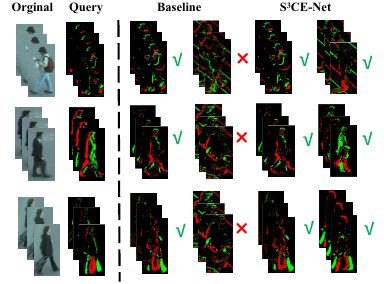}
\caption{Visualization of the retrieval results using Baseline and S$^3$CE-Net. The first and second columns represent the RGB and temporal images of the queried person under camera 1, the third and fourth columns respectively display the retrieval results from the baseline model under camera 1 and camera 2, and the fifth and sixth columns respectively display the retrieval results from S$^3$CE-Net under camera 1 and camera 2. Correct matches and incorrect matches are indicated by check marks and crosses, respectively.}
\label{fig:6}
\end{figure}

\subsubsection{Visualization of Retrieval Results:} 
In Fig ~\ref{fig:6}, we visualize the retrieval results of our S$^3$CE-Net compared to Baseline. 
In some cross-scene person retrieval tasks, our model successfully retrieves video sequences of the same person in different scenes, whereas the baseline  retrieves incorrect cross-scene person sequences. This comparison highlights the superior performance of S$^3$CE-Net.

\begin{table}[h]\small
  \centering
  \caption{Single-domain vs. Cross-domain evaluations (mAP and performance drop) on the blurred dataset, across Event and RGB modalities.  ``Ref. Value'' (Reference Value) represents the model's training and testing on the target dataset.}
  \label{tab:cross_domain_blurred1}
  \setlength{\tabcolsep}{1.5mm}
  \resizebox{\linewidth}{!}{%
  \begin{tabular}{c|c|c}
    \hline
    \textbf{Modality} & \textbf{Training Dataset} 
    & \textbf{Test Dataset \& Performance} \\
    \hline
    \textbf{Event} & \textbf{PRID-2011} & 
    \begin{tabular}{l|c|c}
      \textbf{Blurry} & \textbf{mAP} & \textbf{Drop} \\
      \hline
      Ref. Value & 70.4 & --     \\
      iLIDS-VID & 64.4 & -6.0  \\
    \end{tabular} \\
    \hline
    \textbf{Event} & \textbf{iLIDS-VID} & 
    \begin{tabular}{l|c|c}
      \textbf{Blurry} & \textbf{mAP} & \textbf{Drop} \\
      \hline
      Ref. Value & 90.1 & --     \\
      PRID-2011 & 83.2 & -6.9  \\
    \end{tabular} \\
    \hline
    \textbf{RGB} & \textbf{PRID-2011} & 
    \begin{tabular}{l|c|c}
      \textbf{Blurry} & \textbf{mAP} & \textbf{Drop} \\
      \hline
      Ref. Value & 62.2 & --     \\
      iLIDS-VID & 51.5 & -10.7  \\
    \end{tabular} \\
    \hline
    \textbf{RGB} & \textbf{iLIDS-VID} & 
    \begin{tabular}{l|c|c}
      \textbf{Blurry} & \textbf{mAP} & \textbf{Drop} \\
      \hline
      Ref. Value & 79.7 & --     \\
      PRID-2011 & 67.4 & -12.3  \\
    \end{tabular} \\
    \hline
  \end{tabular}%
  }
\end{table}

\subsection{Domain  Generalization Evaluation}


In this part,  we compare the domain generalization performance of our S$^3$CE-Net in the RGB modality and the event modality.
In addition, we further blur the unknown target domain data to simulate blur caused by high-speed motion or camera shake. This further increases the difficulty of model domain generalization.
\begin{enumerate}[i.]
\item Train the model on the standard PRID dataset and tested it on the unseen blurred iLIDS-VID dataset.
\item Train the model on the standard iLIDS-VID dataset and tested it on the unseen blurred PRID dataset.
\end{enumerate}
As  shown in Table~\ref{tab:cross_domain_blurred1}, compared to single-domain testing, the performance of the model in  the domain generalization task decreases in both RGB and Event modalities. However, compared to the RGB modality, the model is trained on event data, resulting in higher performance and less performance degradation during domain generalization.  This also verifies that compared to the RGB modality, the event modality is beneficial for improving the generalization performance of the Re-ID domain.
This is because RGB data often contains a large amount of background information and scene style information, which are important obstacles to model domain generalization. The event camera can better shield these two parts of information. 

\section{Conclusion}
In this paper, we proposed a  Spike-guided Spatiotemporal Semantic Coupling and Expansion Network (S$^3$CE-Net) for the  long-sequence event Re-ID task.   
Our S$^3$CE-Net  is build on  the  spiking neural networks (SNNs). 
The S$^3$CE-Net contains the  Spike-guided  Spatial-temporal  Attention Mechanism (SSAM) and Spatiotemporal Feature Sampling Strategy (STFS). 
The SSAM conducted the semantic interaction and association in the spatial and temporal  dimensions.  
The  STFS  drived  the Re-ID model's ability to capture robust semantics from  spatiotemporal dimensions. 
Extensive experiments have verified that our S$^3$CE-Net achieves  outstanding performance on many  mainstream long-sequence event  person  Re-ID datasets.

{
    \small
    \bibliographystyle{ieeenat_fullname}
    \bibliography{main}
}
\appendix
\section*{Supplementary Material}
\section{Limitations and Future Work}
Although S$^3$CE-Net demonstrates strong performance, its reliance on spiking backbones introduces latency and hardware constraints. Additionally, our sampling strategy assumes consistent scene dynamics across sub-sequences, which may not hold for highly erratic pedestrian motion. Moreover, the lack of large-scale and realistic event-based Re-ID datasets limits the generalizability and scalability of current methods. Addressing these issues requires more adaptive or motion-aware modules and the construction of more comprehensive benchmarks, which we leave to future investigation.

\section{Dataset Details}
The \textbf{PRID dataset} includes 385 and 749 person identities captured from two different camera viewpoints, with only the first 200 identities appearing in both non-overlapping cameras. This provides a foundation for cross-camera person re-identification. The \textbf{MARS dataset} is a large-scale and diverse dataset that includes 1,261 person identities and 20,008 video sequences, providing a comprehensive benchmark for video-based person re-identification. The \textbf{iLIDS-VID dataset} includes 300 person identities and 600 image sequences captured from two different camera viewpoints.  This dataset is considered challenging due to factors such as similar clothing between different individuals, lighting and viewpoint variations across different camera perspectives, cluttered backgrounds, and random occlusions.

\paragraph{Event-ReID (real-event dataset).}
Event-ReID is the first publicly available \emph{real} event-camera benchmark for person Re-ID.%
\ It was captured by four indoor Prophesee DVS cameras (640\,$\times$\,480) and contains 33 identities
walking across the four views, yielding 16\,k annotated bounding boxes.

\textbf{Pre-processing.}
\begin{enumerate}[label=(\roman*),itemsep=1pt,leftmargin=14pt]
    \item \emph{Spatial cropping.} Each raw event stream is cropped to the pedestrian region using the hand+YOLO bounding box.
    \item \emph{Fixed-time accumulation.} Events are integrated within a constant window $\Delta t = 33.3$\,ms ($\sim$30\,fps), preserving polarity to form one voxel frame per window.
    \item \emph{Long-sequence concatenation.} Consecutive, non-overlapping voxel frames from the same ID--camera tracklet are concatenated, producing clips of 3.4\,s on average and up to 8\,s ($\sim$100--240 frames).
    \item \emph{File organisation.} Each frame is stored as \texttt{IDxxx\_cy\_\#\#\#.txt} under \texttt{output/IDxxx/cam\_y/}; clips are indexed by consecutive numbers.
\end{enumerate}

\textbf{Alternative setting.}
To replicate a constant-event window, run the script with
\texttt{--event\_time=False\ --event\_count=5000} ($\sim$30\,fps); all other steps remain unchanged.

\section{Intra-Frame Relative Position Bias}
To enhance intra-frame attention modeling in SSAM (Spike-guided Spatio-temporal Attention Mechanism), we introduce a learnable intra-frame relative position bias $B$, which captures spatial dependencies among tokens within a single event frame. Given an event tensor with $N_{s} = H_s \times W_s$ tokens, each token $i$ is assigned a 2D coordinate $(h_i,w_i)$. The relative position index between two tokens $i$ and $j$ is defined as:
\begin{equation}
\Delta_{ij} = (h_i - h_j, w_i - w_j)
\end{equation}
Based on this, we construct a learnable relative position bias table $P$ of size:
\begin{equation}
P \in R^{(2H_s -1) \times (2W_s -1)}
\end{equation}
where each entry stores the bias corresponding to a specific spatial displacement. The final relative position bias matrix is indexed as:
\begin{equation}
B_{i,j} = P[relative\_index(\Delta_{ij})],  i, j \in \{1, 2, \dots, N_s\},
\end{equation}
where $relative\_index(\Delta_{ij})$ is used to map spatial displacements to their corresponding entries in the relative position bias table $P$. The design is inspired by Swin Transformer \cite{Liu2021SwinTH}.

 \section{Spiking Neural Networks (SNNs)}
As a third-generation neural network, Spiking Neural Networks (SNNs) differ from traditional neural networks in that they perform asynchronous computation, rather than processing all points globally. 
Additionally, the decay mechanism during the "firing" process allows SNNs to effectively handle temporal information. 
In this work, we adopt Leaky Integrate-and-Fire (LIF) neurons \cite{10.1007/s00422-006-0068-6} to process the complex dynamic features in event data.
The membrane potential update equation for the LIF neuron at layer $l$ at time $t$: 
\begin{equation}
       V_{l}(t+\Delta t)=V_{l}(t)+\frac{\Delta t}{\tau_{m}}\left[-\left(V_{l}(t)-V_{rest} \right)+X_{l}(t)\right]
\end{equation}
\(V_{l}(t)\) is the membrane potential at the current time step $t$, \(V_{rest}\) represents the resting potential, \(\tau_{m}\) is the membrane time constant, and \(X_{l}(t)\) represents the presynaptic input, which is computed as the weighted sum of the pre-synaptic inputs before the spike:
\begin{equation}
    X_l(t)=\sum_{i=1}^n(w_i\sum_j\psi_i(t-t_j))
\end{equation}
\(\psi_i(t-t_j)\) represents the i-th spike event that occurs at time \(t_j\). During the forward propagation process, if \(V_{l}(t) > V_{th}\), a spike is triggered, and the neuron sends a spike signal to the downstream neurons. \(V_{th}\) represents the membrane potential threshold. Voltage reset:
\begin{equation}
V(t)\to V_{reset}
\end{equation}
When the membrane potential \(V_{l}(t) < V_{th}\), the neuron will not fire a spike. The membrane potential \(V_{l}(t)\) gradually accumulates until it exceeds the threshold or continues to remain below the threshold.

\end{document}